# A Formal Look at Dependency Grammars and Phrase-Structure Grammars, with Special Consideration of Word-Order Phenomena[*]

Owen Rambow and Aravind Joshi

Department of Computer and Information Science

University of Pennsylvania

Philadelphia, PA 19103, USA

rambow,joshi@linc.cis.upenn.edu

**Abstract**

The central role of the lexicon in Meaning-Text Theory (MTT) and other dependency-based linguistic theories cannot be replicated in linguistic theories based on context-free grammars (CFGs). We describe Tree Adjoining Grammar (TAG) as a system that arises naturally in the process of lexicalizing CFGs. A TAG grammar can therefore be compared directly to an Meaning-Text Model (MTM). We illustrate this point by discussing the computational complexity of certain non-projective constructions, and suggest a way of incorporating locality of word-order definitions into the Surface-Syntactic Component of MTT.

## 1 Introduction

In the past, there has been little contact between those linguistic traditions working on the basis of dependency grammars (DGs), such as Meaning-Text Theory (MTT), and those linguistic traditions working on the basis of context-free phrase-structure grammars (CFGs), such as the various incarnations of Chomskyan Transformational Grammar (which we will henceforth refer to by a recent name, Government and Binding Theory or GB). The linguistic insights from one tradition have generally not been transferred to the other. While this fact has been discussed in the past (see, e.g., (Nichols, 1979)), the discussion has not specifically addressed the effect of the underlying formalisms on the linguistic theories that are developed in them. This is not to say that the formalisms themselves have not been compared (see (Gaifman, 1965)); however, while both MTT and GB developed out of formal and/or computational approaches, both theories have shed their original explicitly mathematical underpinnings. A mathematical comparison between the underlying formal systems will therefore not tell us much about the linguistics of the two theories. Instead, we must ask how the formalisms affect the linguistic theories which are expressed in them.

[*] A version of this paper was presented at the Workshop on the Meaning-Text Theory, Darmstadt, Germany, August 1992. The Proceedings are available as "Arbeitspapiere der GMD 671". The authors would like to thank Richard Hudson and three anonymous reviewers for very helpful and insightful comments and suggestions. This work was partially supported by the following grants: ARO DAAL 03-89-C-0031; DARPA N00014-90-J-1863; NSF IRI 90-16592; and Ben Franklin 91S.3078C-1.

The formalisms that linguistic theories use for the purpose of expressing syntactic structure can differ in two ways. Firstly, the formalisms can differ in the type of representation they use. A phrase-structure grammar postulates the existence of non-terminal syntactic categories, while a dependency grammar does not. Secondly, the linguistic theories can differ in how they use the syntactic formalism they have chosen. Early Chomskyan approaches followed a generativist approach, while MTT does not. As has been pointed out previously (Kunze, 1972, p.10), these two issues – the definition of the formalism itself and how it is used – are orthogonal. It is perfectly possible to define a generative DG; see for instance (Hays, 1964). While the difference between a generativist and a non-generativist approach will have profound methodological (and perhaps philosophical) implications for the resulting linguistic theories, in this paper we will concentrate on the representational difference in the formalisms themselves.

It has often been observed that a key linguistic difference between MTT (and other dependency-based theories) on the one hand and GB on the other hand is the central role that the lexicon plays in MTT, but not in GB (see, e.g., (Sgall, 1992)). We claim that this difference is not coincidental, but a mathematical property of the underlying formalisms. Context-free phrase-structure grammars *cannot* be the basis of a lexicon-oriented linguistic theory (in a technical sense, which we will define in the next section), while dependency grammars *must* be[1]. An attempt to "lexicalize" CFGs leads naturally to a more powerful phrase-structure system called Tree Adjoining Grammars (TAGs)[2]. In Section 2, we will show that TAGs show many important similarities to DGs. These similarities have two beneficial results: firstly, we are able to apply formal results from the mathematical study of phrase-structure grammars to DGs; secondly, we are able to transfer linguistic analyses made in one framework to the other framework. In Section 3, we will illustrate these points by looking at several non-projective syntactic constructions.

Our major goal in this paper is to study the interaction between formal systems and linguistic theories, and to explore how results in the framework of one theory can be expressed in the particular formal context of other theories. Such work provides insights into those aspects, linguistic and formal, that appear to be invariant across a class of formalisms. The reader should not interpret our goal as suggesting that MTT needs to adopt a phrase-structure representation for whatever reason!

## 2  Dependency, Phrase Structure and the Lexicon

One of the most important features of MTT is the central role that the lexicon plays (see e.g. (Mel'čuk and Polguère, 1987); in fact, much of the MTT literature deals with the lexicon). For syntactic purposes, it contains information about the subcategorization frame of a lexeme, and how the arguments are realized (case assignment and function words). The importance of the lexicon for syntactic theories has also been increasingly recognized in the American linguistic traditions. We will take it as a given, and address the question how a phrase structure-based syntactic theory can be adapted to a lexical approach. It turns out that there are intrinsic, formal problems. These problems have been investigated in detail by (Schabes, 1989); for a summary of some of the mathematical properties of tree grammars including lexicalization, see (Joshi and Schabes, 1991). We will provide a brief discussion here.

---

[1] From a historical perspective, it presumably was the interest in developing a lexicon-oriented linguistic theory that led to the use of a dependency grammar for MTT. However, in this paper we take a synchronic view.

[2] TAG was originally introduced as a tree generating system on its own (Joshi et al., 1975). It was only recently shown that TAGs can lexicalize CFGs. In this paper, we will only be interested in lexicalized TAGs. For a general introduction to TAGs, see (Joshi, 1987a).

If we want to analyze how formal systems can be used for linguistic theories, we must start by determining what sort of *elementary structures* the formalism provides, and how these elementary structures are combined using the *combining operations* defined in the formalism. We will illustrate these notions with some examples. First, consider CFGs. In a CFG, a grammar consists of a set of rewrite rules, which associate a single nonterminal symbol with a string of terminal and nonterminal symbols. Here is a sample context-free grammar:

(1) a. S ⟶ NP VP

b. VP ⟶ really VP

c. VP ⟶ V NP

d. V ⟶ likes

e. NP ⟶ John

f. NP ⟶ Lyn

Each of these rules is an elementary structure in this grammar. We combine these elementary structures by using one rule to rewrite a symbol introduced by another rule. For example, when we use rule (1a), we introduce the nonterminal symbols (or "nodes") NP and VP. We may rewrite the VP node by using rule (1b) or (1c). This grammar generates, among others, the following string:

(2) John really likes Lyn

Derivations in CFGs can be represented as trees: for each nonterminal node in the tree, the daughters record which rule was used to rewrite it. The phrase-structure tree that corresponds to sentence (2) is given in Figure 1.

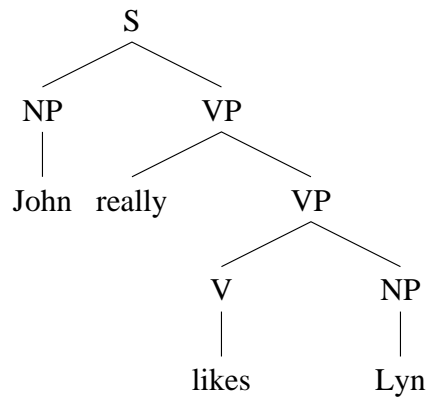

Figure 1: Phrase Structure Tree for *John really likes Lyn*

Now consider a different type of mathematical formalism, Tree Substitution Grammars (TSG). In a TSG, the elementary structures are phrase-structure trees. A sample grammar is given in Figure 2. It consists of three trees, one of which is rooted in S, and two of which are rooted in NP. Note that even though from the point of view of a CFG, a tree is a derived object, not an elementary one, we have defined TSGs in such a way that a tree is now an elementary object of the grammar.

We combine elementary structures in a TSG by using the operation of *substitution*, illustrated schematically in Figure 3. We can substitute tree $\beta$ into tree $\alpha$ if there is a nonterminal symbol on the frontier of $\alpha$ which has the same label as the root node of $\beta$ ('A' in Figure 3). We can then simply append $\beta$ to $\alpha$ at that node. (Nodes at which substitution is possible are called "substitution nodes" and are marked with down-arrows ($\downarrow$).) A derivation in our sample TSG is shown in Figure 4. The trees representing the two arguments of the verb *like*, *John* ($\alpha_2$) and *Lyn* ($\alpha_3$), are substituted into the tree associated with the verb ($\alpha_1$), yielding the well-formed tree $\alpha_4$, from which the sentence *John likes Lyn* can be read off.

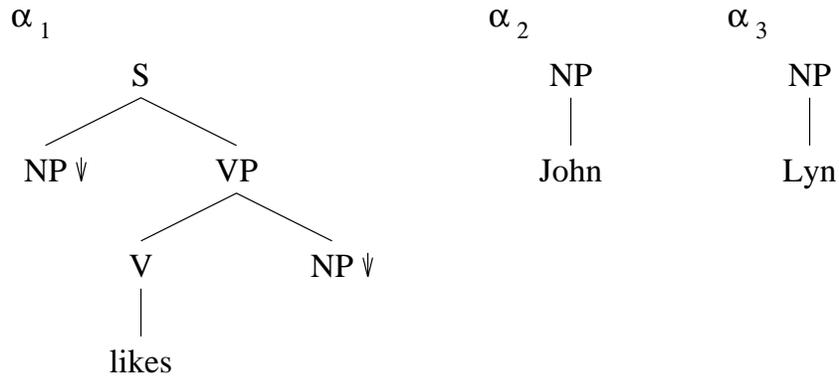

Figure 2: A sample TSG

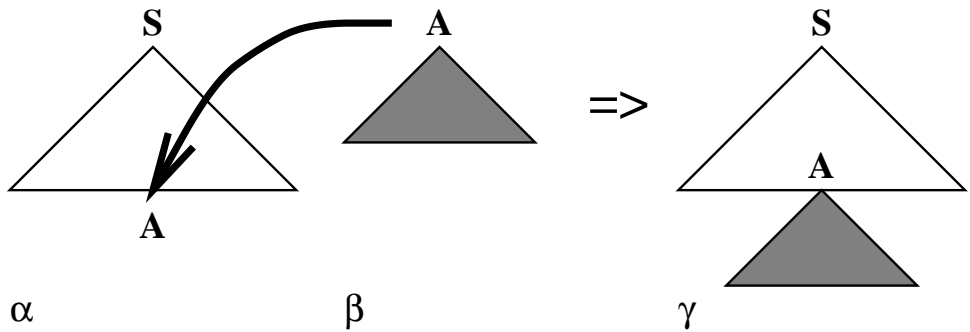

Figure 3: The Substitution Operation

Finally, compare the DG used for MTT to CFG and TSG. In DG, the elementary structures are simply nodes labeled with terminal symbols, i.e., lexical items. There are no nonterminal symbols. Nodes are composed by establishing dependency relations between them. The result is a dependency tree.

CFGs and TSGs are weakly equivalent (they generate the same languages). However, to a linguist, they look very different. A context-free rule contains a phrase-structure node and its daughters; an elementary tree in a TSG may be of arbitrary height. Put differently, we have increased the "domain of locality" of the elementary structures of the grammar. This increased domain of locality allows the linguist to state linguistic relationships (such as subcategorization, case assignment and agreement) differently in a TSG. As an example, take agreement between subject and verb in English. The linguist working in TSG can simply state (by using some feature-based notation) that the verb and the NP in subject position in tree $\alpha_1$ of Figure 2 agree with respect to a given set of features. The linguist working in CFG has a harder time: since the verb is in rule (1d), while the subject NP is in rule (1a), he cannot simply state the relation directly, since it is impossible to state constraints that relate nodes in different elementary structures. Instead, the linguist must propose that the NP in fact agrees with the VP in rule (1a), and that the VP agreement features are inherited by its head in rule (1d). The notion that the VP (and not only the verb) agrees with the subject is a meaningful linguistic proposition, and in fact the TSG linguist could have adopted it as well. However, the crucial issue is that the CFG linguist, because of his choice of formalism, was *forced* to adopt it, while the TSG linguist may choose to do so or not on independent grounds.

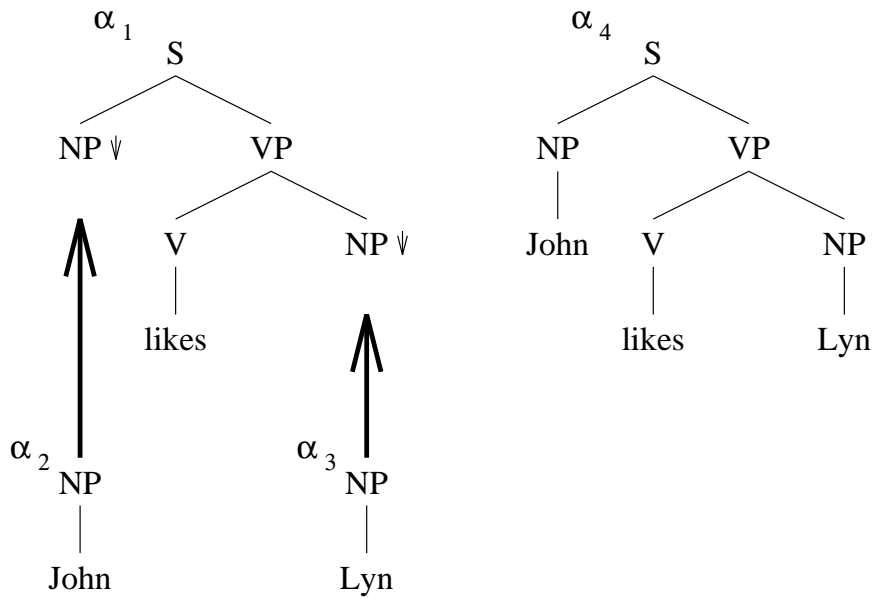

Figure 4: Substitution of arguments into initial tree of *likes*

Now let us turn to our central concern, the role of the lexicon. We will call a grammar "lexicalized" if every elementary structure is associated with exactly one lexical item, and if every lexical item of the language is associated with a finite set of elementary structures in the grammar. Clearly, dependency grammars (including MTT) are naturally lexicalized in this sense, since the elementary structures simply are the lexical items.

The case is more complex for a CFG. Consider the sample grammar given above in (1). We can see that no lexical item is associated with rules 2a and 2c; therefore, the grammar is not lexicalized. It would be possible to combine rules 2a, 2c and 2d into a single one:

(3) S $\longrightarrow$ NP likes NP

However, it is now impossible to correctly place the adverb *really*, since a (lexicalized) rule of the form (1b) is no longer useful (the VP node having been eliminated). The adverb cannot be inserted between the subject and the verb.

There is a second way of lexicalizing a CFG: instead of merging the two phrase-structure rules into a single rewrite rule, we can combine them and consider the result – a fragment of a phrase structure tree – an elementary structure. Put differently, we move from CFG to TSG. For example, tree $\alpha_1$ in Figure 2 is the result of combining rules (1a), (1c) and (1d). As desired, tree $\alpha_1$ is associated with exactly one lexical item, the verb *likes*. Thus, we have now obtained a TSG from a CFG. We can derive the sentence *John likes Lyn* as shown previously in Figure 4.

It turns out that a TSG is not really what we want, either: we are again faced with the problem of getting the adverb in the right place, since there is no node into which to substitute it[3]. This problem is solved by the tree composition operation of *adjunction*, introduced in the framework of Tree Adjoining Grammars (TAG). Adjunction is shown in Figure 5. Tree $\alpha$ (called an "initial tree") contains a non-terminal node labeled $A$; the root node of tree $\beta$ (an "auxiliary tree") is also labeled $A$, as is exactly one non-terminal node on its frontier (the "foot node"). All other frontier nodes are terminal nodes or substitution nodes. We take tree $\alpha$ and remove the subtree rooted at its node $A$, insert in its stead tree $\beta$, and then add at the footnode of $\beta$ the subtree of $\alpha$ that we removed earlier. The result is tree $\gamma$. As we can see, adjunction can have the effect of inserting one tree into the center of another. Our linguistic example is continued in Figure 6. Tree $\beta_1$ containing the adverb is adjoined at the VP node into tree $\alpha_4$. The result is tree $\alpha_5$, which corresponds to sentence (2). Note that $\alpha_5$ is composed of trees $\alpha_1$, $\alpha_2$, $\alpha_3$ and $\beta_1$, each of which correspond to exactly one lexical item, in contrast to the grammar given above in (1).

---

[3] Having two verbs *like*, one of which also subcategorizes for an adverb, does not solve the problem, since it does not generalize to multiple adverbs (in addition to being linguistically unappealing).

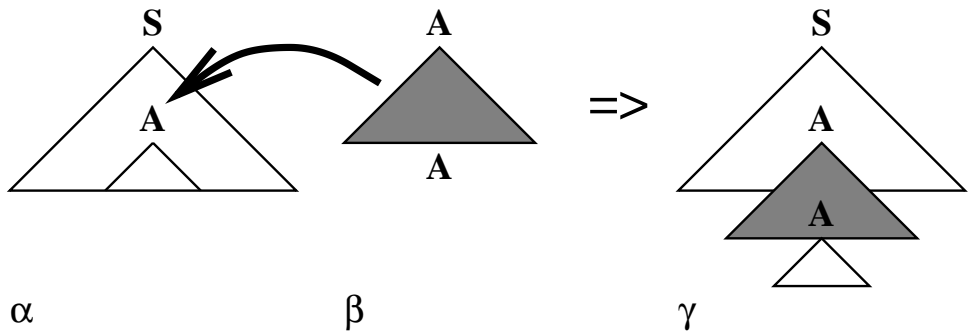

Figure 5: The Adjunction Operation

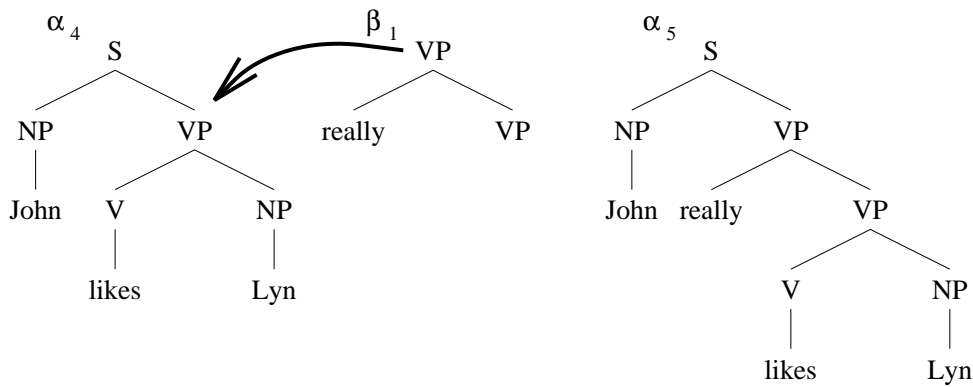

Figure 6: Adjunction of *really* into initial tree

A formalism in which the elementary structures of a grammar are trees and in which the combining operations are adjunction and substitution is called a TAG. (Schabes, 1989) has shown that a tree composition system is only lexicalizable if the composition operations include adjunction. Thus, the process of lexicalizing a CFG naturally leads to a TAG. TAGs are more powerful formally than CFGs, meaning that they can derive more complex languages than CFG. They are also more difficult to parse. Several other proposals have been made to adapt phrase-structure grammars to a lexical approach, including categorial formalisms such as CCG (Steedman, 1991), and non-transformational phrase structure grammars such as LFG (Bresnan and Kaplan, 1982) and HPSG (Pollard and Sag, 1987). Interestingly, the underlying formalisms of these frameworks are also more powerful than CFG. For a summary of mathematical and computational properties of TAGs and some related phrase-structure formalisms, see (Joshi et al., 1991).

Like a CFG, a TAG derives a phrase-structure tree, called the "derived tree". (The derived tree for our example is the right tree in Figure 6.) In addition to the derived (phrase-structure) tree, a second structure is built up, the "derivation tree". In this structure, each of the elementary trees is represented by a single node. Since the grammar is lexicalized, we can identify this node with the (base form of the) lexeme of the corresponding tree[4]. If a tree $t_1$ is substituted or adjoined into a tree $t_2$, then the node representing $t_1$ becomes a dependent of the node representing $t_2$ in the derivation tree. Furthermore, the arcs between nodes are annotated with the position in the "target tree" at which substitution or adjunction takes place. In the TAG literature, this annotation is in the form of the tree address of the node (using a formal notation to uniquely identify nodes in trees, without reference to linguistic concepts). However, in analogy to the MTT notation, we can simply assign numbers to argument positions, and introduce the convention that all other positions are attribute positions, marked as ATTR. The derivation tree for the example derivation above is shown in Figure 7. We can see that the derivation structure is a dependency tree which closely resembles the Deep-Syntactic Representation (DSyntR) of MTT.

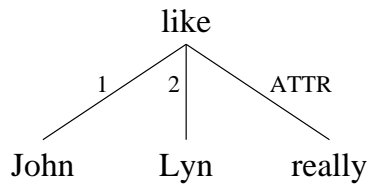

Figure 7: Derivation Tree for *John really likes Lyn*

The resemblance between the derivation structure and the DSyntR is not a coincidence. It is a direct result of lexicalization. We would like to summarize some striking similarities between an MTM of a language and a TAG grammar for that language.

1. As in the case of an MTM, a grammar in the TAG formalism consists of a lexicon, whose elementary structures are combined by some very simple rules of composition (substitution and adjunction in the case of TAG).

2. The function words are included in the elementary structures of the lexemes that require them in their subcategorization frame, i.e., they are represented in the lexical entries for content words, not separately. They are therefore not represented in the derivation structure, just as they are not represented in the DSyntR.

3. A verb subcategorizes for its arguments – there must be exactly one constituent for each of its obligatory arguments. Adjuncts are not subcategorized for and there is no (syntactic) limit on their number. In MTT, this is reflected by the fact that there may an unbounded number of ATTR subtrees, while there is only one subtree for each of the numeric arc labels. In TAG, this distinction is captured by the fact that arguments are substituted, a unique and obligatory step, while adjuncts are adjoined, a recursive but optional step.

4. In a TAG, the lexicon consists of one tree family for each lexeme, each tree family containing trees for the syntactic variants of the lexeme (active/passive voice, *wh*-questions for each argument, topicalization for each argument etc.). As in the case of MTT, certain syntactic paraphrases can be handled by general rules ("metarules", (Becker, 1990)). Lexical functions and syntactic paraphrases that use lexical functions have not yet been introduced in the TAG framework, but they could be integrated in a straightforward way.

---
[4]This is not exactly what is done in the TAG literature, but the difference is purely notational.

5. Idioms (phrasemes) have been discussed both within the TAG framework (Abeillé and Schabes, 1989) and in MTT (Mel'čuk, 1988, p.60). Both frameworks can account for idioms in a natural and similar way, namely by postulating elementary structures that (non-compositionally) contain more than one lexeme.

However, it is important to note some important differences between the two approaches:

1. In TAG, word order must be determined at the same time as dependency. This process cannot be separated into two steps, as in MTT[5]. This means that the lexicon in a TAG grammar for a specific language must contain more syntactic information than a lexicon in the MTT framework: not only must it contain information about subcategorization and function words, the trees themselves must also contain enough information so that the word order comes out right.

2. While substitution of a tree $t_1$ into tree $t_2$ corresponds to a dependency of the lexemic element of $t_1$ on that of $t_2$, this need not be the case in adjunction. We will see later examples in which $t_1$ is adjoined into $t_2$, but the lexemic element of $t_2$ depends on that of $t_1$. Thus, while adjunction corresponds to the establishment of a syntactic dependency relation, the direction of the relation cannot be determined from the direction of the adjunction alone.

The similarities between MTT and a TAG approach, both in the linguistic approach and in the resulting representations, allow us to use TAG as a way of relating MTT analyses to phrase-structure-based analyses. While much of the work on the interface between syntax and semantics, on lexical functions and on syntactic paraphrases in the MTT framework can be reformulated in terms of a TAG analysis, we will concentrate in this paper on applying insights from TAG analyses to the MTT framework.

## 3   Formal Aspects of Word Order Variation

In this section, we will use the close relationship between lexicalized TAGs and DGs to make some observations about non-projective constructions. There are two potential problems with non-projective[6] constructions in a dependency-based theory:

- No parsing model for non-projective constructions is known that is computationally "well-behaved".

- The syntax of non-projective constructions must be expressed differently from that of projective constructions, which is linguistically unmotivated.

We discuss the first point in more detail in Section 3.1. Sections 3.2 through 3.4 discuss three illustrative syntactic constructions. We address the second point in Section 3.5, where we present a proposal for handling certain non-projectivity within the MTT notion of "syntagm".

---

[5]There have been proposals for formal variants of TAG in which the linear precedence of nodes is stated independently from immediate dominance; see (Joshi, 1987b; Becker et al., 1991).

[6]We take the standard definition of projectivity, as given in (Mel'čuk, 1988, p.35f), which can be shown to be equivalent to the definitions discussed in (Marcus, 1965).

## 3.1 Computational Properties of Dependency Grammars

The principal reason for studying mathematical aspects of the syntactic formalism used by a linguistic theory is probably the need to explain the computational processes involved in the generation and understanding of language. While it appears that most syntactic constructions in most languages are projective (Mel'čuk and Pertsov, 1987, p.184), many languages do have syntactic constructions (often, but not always, pragmatically marked) that are not. It has been shown that a fully projective dependency grammar is weakly equivalent to a CFG (Gaifman, 1965), where "weak equivalence" means that for every DG, there is a CFG that generates exactly the same set of sentences, and *v.v.* The equivalence of projective DGs and CFGs lets us transfer parsing results from CFGs to such grammars. In particular, we know that we can parse a string in a CFG in at most $O(n^3)$ time, i.e. in an amount of time proportional to the cube of the length of the input string. Though the parsing of non-projective DGs has been discussed (see (Covington, 1990) and the references therein), to our knowledge no formal result has been published. There is reason to believe that in the worst case they can be parsed in a time proportional to an exponential function of the length of the input string ($O(2^n)$). If this worst case actually occurred in natural language parsing, then a DG would not be a very appealing candidate for a model of human language processing.

Why is this a potential problem for MTT? Humans appear to be quite good at parsing, i.e. constructing a syntactic representation for a linear string of language. If a linguistic theory wants to account for this process, then it must be able to provide an account of how the syntactic structures the theory postulates can be effectively and efficiently constructed from the input. Even if a linguistic theory does not aim at providing an account of human sentence processing (as in fact neither MTT nor GB do), then it must be the case that such an account can, in principle, be found, since otherwise the relation of the theory to observable behavior is unclear. An account of human sentence processing must be inherently computational. While a mathematical study cannot, of course, provide a computational theory of processing, it can provide useful guidelines for the elaboration of such a theory, and thus confirm the possibility of elaborating such a theory.

Of course, it could be argued that non-projective constructions are in fact much more difficult to process than projective ones, and that therefore the lack of a processing account for non-projective trees is actually welcome, rather than a problem. However, data from psycho-linguistic experiments suggests that processing difficulty does not pattern with the projective/non-projective distinction (or, equivalently, the distinction between CFGs and more powerful formalisms). For example, Bach et al. (1986) show that the non-projective Dutch cross serial dependencies (which we discuss in Section 3.3 below) are in fact easier to process than German projective nested dependencies. Joshi (1990) gives a TAG-based account of these differences that crucially relies on the fact that both constructions can be handled by the same mathematical formalism.

In the previous section we have argued that analyses using lexicalized TAG and dependency-based analyses bear striking resemblances. In these sections, we will exploit these resemblances and discuss two types of deep non-projectivity (i.e., non-projectivity which affects lexical items that are already present at the level of DSyntR). We will argue that the non-projectivity caused by *wh*-words in embedded English sentences and by the Dutch cross-serial dependencies can be handled by a TAG. Since TAGs can be parsed in $O(n^6)$ time (Vijay-Shanker, 1987), we can conclude that these types of non-projectivity are "well behaved" from a processing point of view. We then briefly discuss a third type of non-projectivity, scrambling in German, which cannot be handled by a TAG, either.

## 3.2 Embedded *wh*-words in English

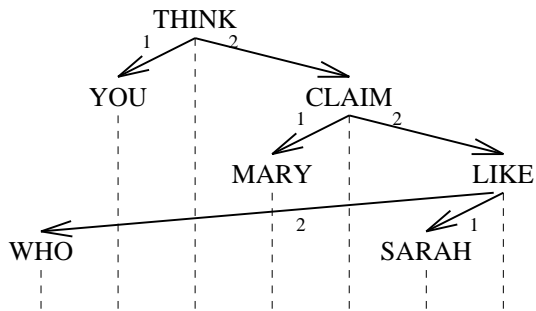

Figure 8: DSyntR for Sentence (4)

Like in many other languages, *wh*-words in English questions generally must appear in sentence-initial position. This is also true in the case that the *wh*-word is an argument of an embedded verb. Strikingly, there is no bound on the depth of the embedding:

(4) Who do you think that Mary claimed that Sarah liked?

In (4), the *wh*-word is an argument of the most deeply embedded verb *like*, thus causing the non-projectivity, as can be seen in Figure 8. A TAG can capture the long-distance dependency naturally, since the recursive adjunction operation allows an unbounded number of clauses to intervene between directly dependent lexemes. An analysis of *wh*-movement in the TAG framework has been proposed by Kroch (1987); our analysis (Figure 9) is a slight variation of his analysis. The steps are shown in Figure 9. We first substitute all nominal arguments into their respective verbal trees, and then adjoin the intermediate *claim*-clause ($\beta_2$) into the most deeply embedded *like*-clause ($\alpha_1$) at the S node immediately domainted by the root. This has the effect of separating the *wh*-word from its verb, even though they originated in the same structure. We then subsequently adjoin the matrix *think*-clause ($\beta_1$) into the intermediate *claims* clause.

The derivation leads to two structures: the derivation tree in Figure 10, and the derived tree in Figure 11. The derivation structure records the sequence of adjunctions and substitutions that leads to the derived tree, while the derived tree in Figure 11 shows the phrase structure and thus the word order of the final sentence. These two structures exist in parallel; we do not have to determine the word order from the dependency-based derivation tree as a separate step.

The reader will observe that, contrary to the example of sentence (2), the derivation structure (given in Figure 7) does not correspond directly to the DSyntR: the direction of adjunction between the verbs (more precisely, the trees anchored in verbs) does not correspond to the direction of the dependency.[7] Why is this? We have seen that nominal arguments are substituted into verbal trees, and that adjuncts are adjoined into trees they modify. In both instances, the derivation structure corresponds to the dependency structure. However, in the analysis for embedded clauses we have given here, we adjoin the matrix clause into the dependent clause at its S node. This is indicated in the derivation tree (Figure 10) by annotating the arcs with an 'S' rather than with an MTT-style annotation. This difference, however, does not affect the point we would like to make in this paper: what is central to this exposition is that a derivation in a TAG is like a dependency analysis in that it establishes direct relation between lexical items. The direction of adjunctions need not correspond to the direction of the dependency, as long as the latter can be retrieved from the former by some linguistically motivated simple procedure. For example, in our case, the actual dependency structure can be derived trivially:

---

[7] For many constructions, the exact dependency analysis is often a matter of discussion. However, in the case at hand, the issue is quite uncontroversial.

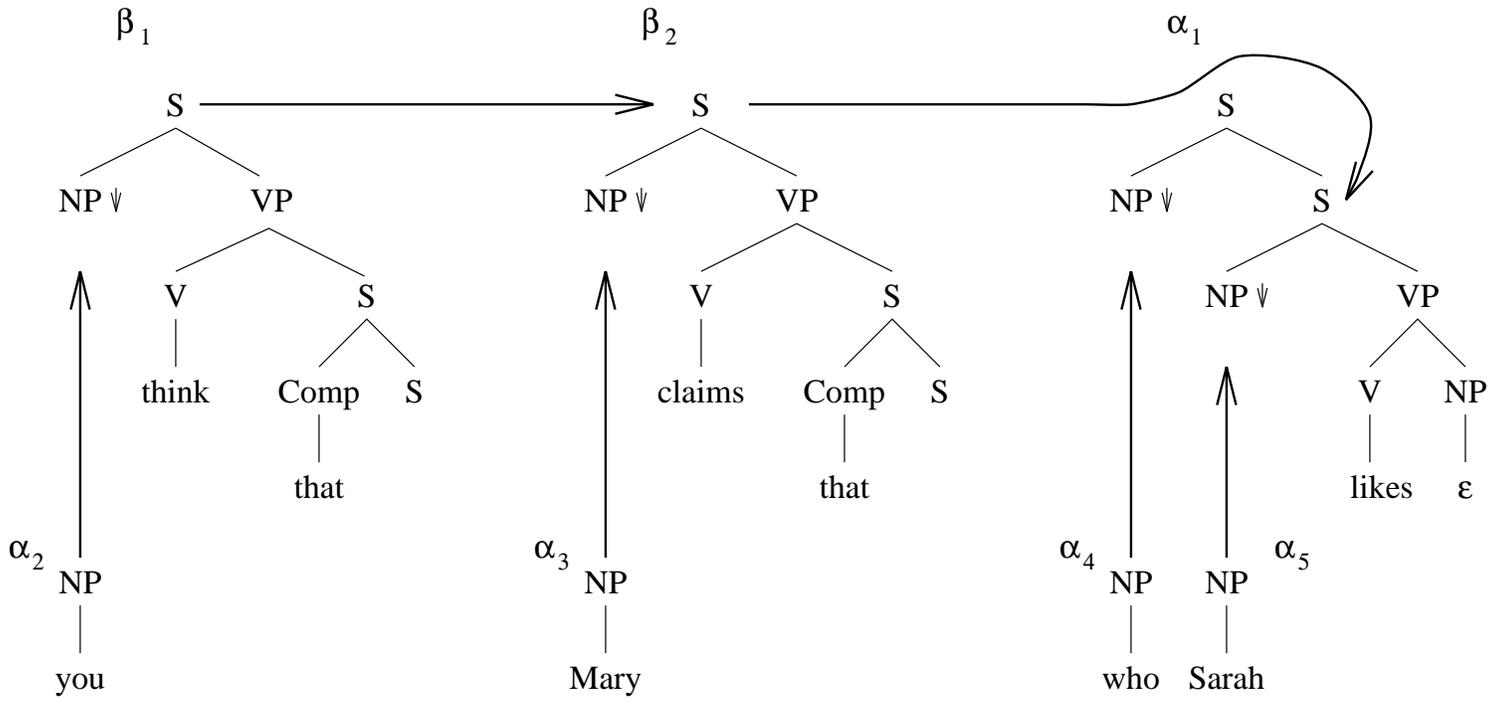

Figure 9: TAG derivation for Sentence (4)

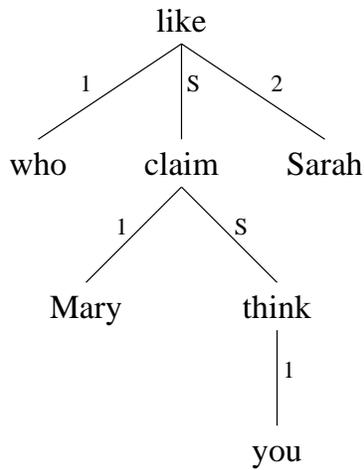

Figure 10: TAG derivation tree for Sentence (4)

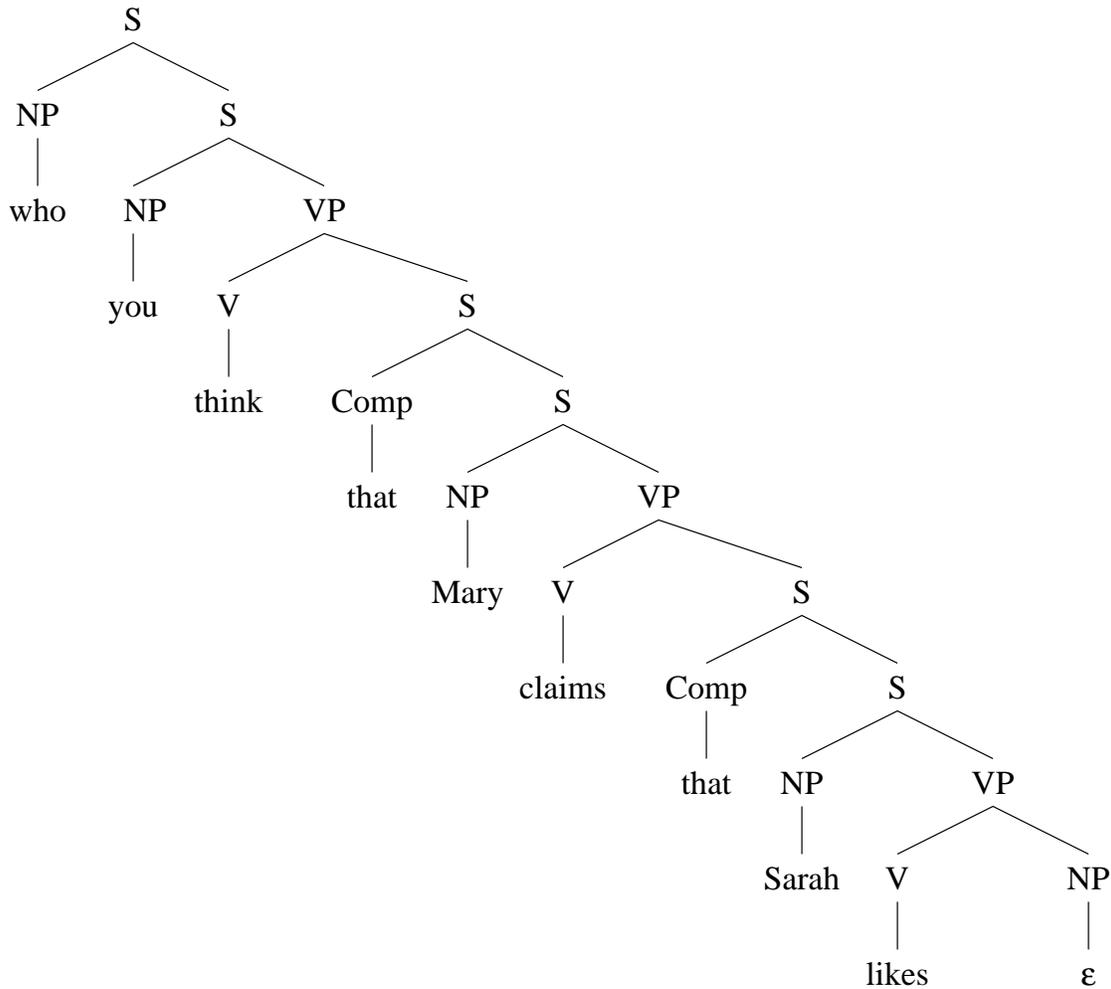

Figure 11: TAG derived tree for Sentence (4)

arcs marked 'S' are simply inverted.

We may draw two conclusions. First, since the construction can be represented by a TAG, we can parse this type of non-projectivity in $O(n^6)$ time. Second, we can state the word-order rules locally in the tree associated with one clause: in tree $\alpha_3$ in Figure 9, the *wh*-word has been moved to the front of the clause. This local operation becomes a non-projective one through adjunction. In Section 3.5 we will propose a way of implementing this locality of word-order rules in the MTT framework.

## 3.3 Embedded Clauses in Dutch

As in German, embedded clauses in Dutch can occur before the (clause-final) verb in a "center-embedding" construction. However, the order of the verbs in the two languages differ: while in German, the dependencies between the verbs and their arguments are nested, they are cross-serial in Dutch. Consider the following sentence:[8]

(5) ...omdat   Wim Jan Marie de kinderen zag helpen leren    zwemmen
    ...because Wim Jan Marie the children  saw to help to teach to swim
    ...because Wim saw Jan help Marie teach the children to swim

---
[8]We would like to thank Hotze Rullmann and Marc Verhagen for helping us with this example.

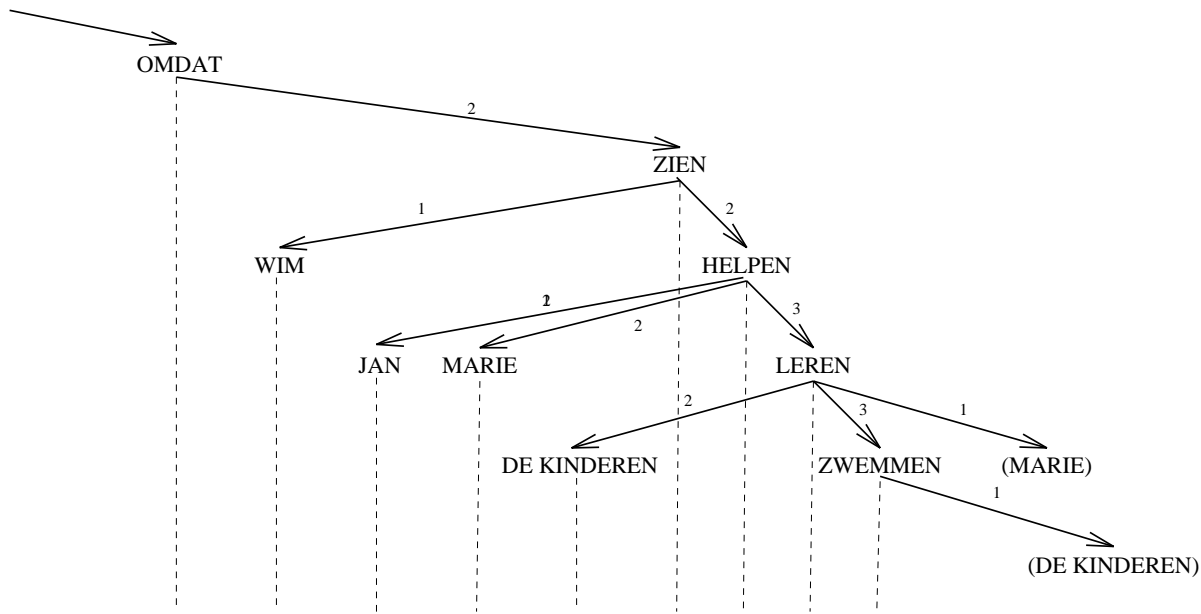

Figure 12: DSyntR for Sentence (5)

This construction is one of the well-known non-projective constructions (see e.g. (Mel'čuk, 1988, p.38)), as can be seen in Figure 12.[9][10] Our TAG analysis in Figure 13 is based on that proposed in (Joshi, 1987a; Kroch and Santorini, 1991). The main verb of each clause is "raised", an analysis proposed independently of the TAG analysis in the GB literature. We then adjoin each clause into its immediately dependent clause at the S node immediately dominated by the root node. This "pushes" both verbs away from their nominal arguments, even though they originate in the same elementary structure. The order of the verbs in the final sentence simply follows from the way the elementary structures are adjoined; no global word-order rules are necessary.

We again conclude that this type of non-projectivity can be parsed in $O(n^6)$ time, and that the word-order rules can be expressed as local constraints on clause-sized structures.

## 3.4 "Scrambling" in German

We see that in the cases of *wh*-elements in embedded clauses in English and of Dutch cross-serial dependencies, TAGs can provide an account. We avoid an exponential explosion in computing time for the parsing problem. However, it appears that there are constructions in natural language that surpass even the additional power of TAGs. One such construction derives from the free word order allowed in verb-final languages such as German, Japanese and Hindi. In German, more than one actant from an embedded clause may be ordered among the actants of the matrix clause. We will refer to these actants as "scrambled". In the matrix clause, the scrambled embedded constituents may occupy any position. An example is given in sentences (6a) and (6b) below.

(6) a. ... daß der Detektiv$_i$    niemandem   [PRO$_i$ den Verdächtigen
       ... that the detective (nom) no-one (dat)      the suspect (acc)

---

[9]In this and other DSyntRs, actants that are deleted at subsequent stages (i.e., at SSyntR) are represented in parentheses. Furthermore, in DSyntR trees we follow the common practice of labeling nodes with the infinitives of verbs, while for TAG trees we will label nodes with the fully inflercted form.

[10]The non-projectivity of the construction is independent of the particular analysis chosen for "control" verbs of various types.

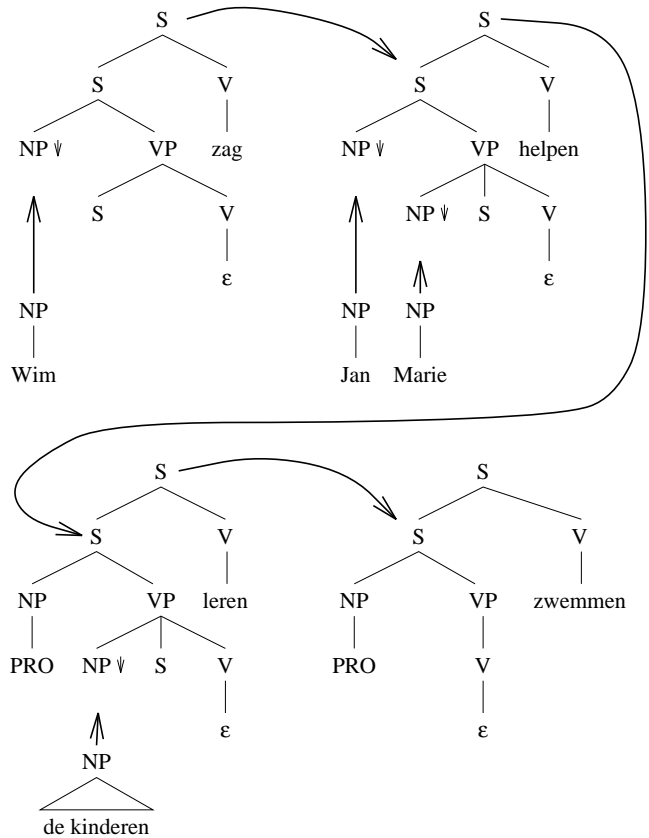

Figure 13: TAG Derivation for Sentence (5)

    des Verbrechens zu überführen] verspricht
    the crime (gen) to indict    promises

    ...that the detective has promised no-one to indict the suspect of the crime

  b. ...daß [des Verbrechens]$_k$ der Detektiv$_i$    [den Verdächtigen]$_j$
    ...that the crime (gen)    the detective (nom) the suspect (acc)

    niemandem  [PRO$_i$ $t_j$ $t_k$ zu überführen] verspricht
    no-one (dat)          to indict    promises

    ...that the detective has promised no-one to indict the suspect of the crime

In this particular sentence, we have a mixture of the cross-serial dependencies of Dutch and of the nested dependencies of standard German. Such constructions pose special problems even for TAGs[11]. In (Becker et al., 1991), we provide a discussion of the issue, and show that a more powerful extension of TAG, "Multi-Component TAG" (MC-TAG), can handle all cases. In MC-TAGs, first introduced by Weir (1988), the elementary tree is split up into parts, which are grouped together into sets. All trees from one set must be adjoined at the same time. The derivation of the example sentence is shown in Figure 15, with the result in Figure 16. The matrix clause is represented by a tree set containing two trees. These trees are adjoined at different nodes into the single tree representing the embedded clause.

While we have an analysis using MC-TAG, this does not help us with the complexity of the parsing problem: the parsing problem for the relevant version of MC-TAG, called "non-local MC-TAG", is known to be NP-complete (see (Rambow and Satta, 1992) for a summary of relevant mathematical and computational properties), which means that with high likelihood it is exponential (and thus no better than the general parsing problem for non-projective DGs). Scrambling remains an open problem from the processing point of view.

---

[11]In the case of one overt nominal argument per clause, TAGs can handle sentences involving 4 or fewer embedded clauses.

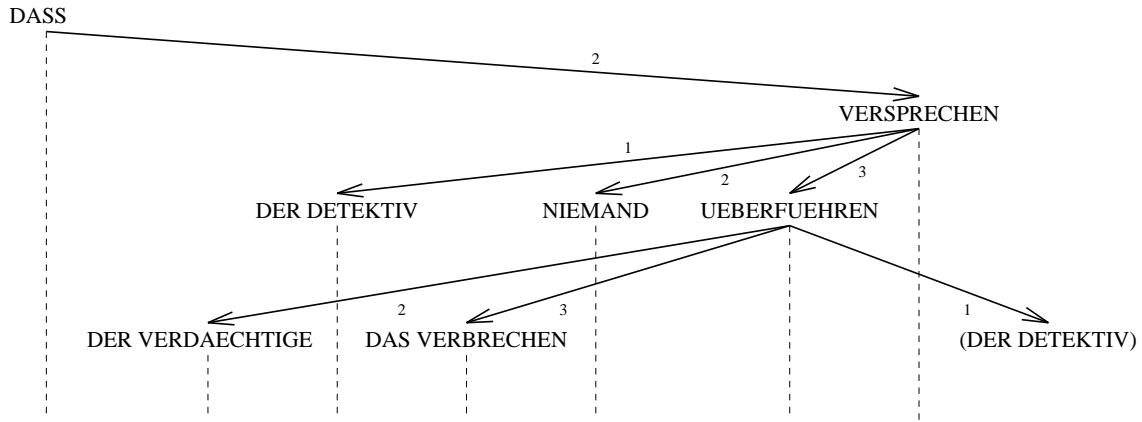

Figure 14: DSyntR of Sentence (6b)

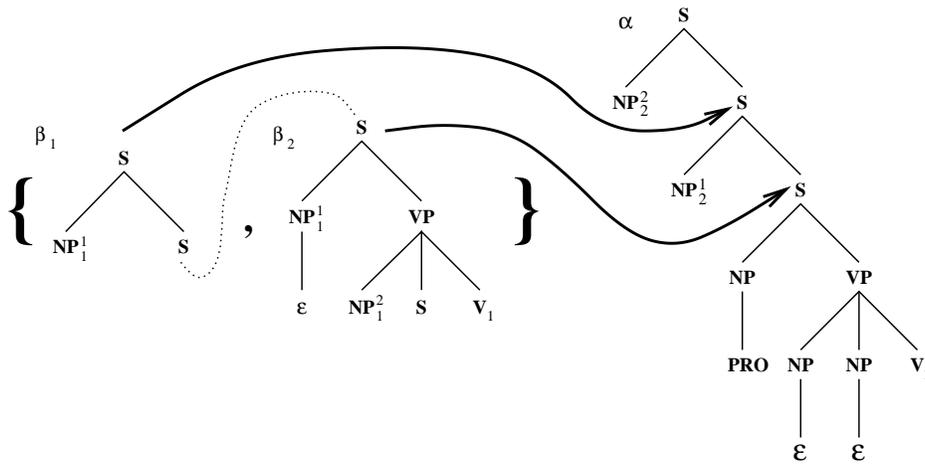

Figure 15: Deriving the Scrambled Sentence (6b)

## 3.5 Localizing Syntactic Rules

As discussed in (Mel'čuk and Pertsov, 1987, p.180), word-order rules for unbounded non-projective constructions cannot be stated as "syntagms", i.e., as local rules affecting two nodes linked by a dependency relation, or as conditions on syntagms. (Here, "unbounded non-projectivity" means that there is no limit on the number of lexical items simultaneously in violation of projectivity.) Instead, they must be stated in separate global rules. The existence of two types of word-order rules is not fully satisfactory: it is motivated not by any linguistic considerations, but only by the mathematical properties of the underlying dependency formalism; and it contradicts the spirit of Mel'čuk's Principle of Maximal Localization (Mel'čuk, 1988, p.387).

As we have seen, in the case of *wh*-words and Dutch embedded clauses, the TAG approach lets us localize the word-order rules within the elementary structure of a clause (or, more precisely, of a verb), just as, say, the SOV order is localized in elementary trees. How can we transfer this approach to MTT? One way of localizing all word-order rules would be to associate phrase-structure trees with nodes of the dependency tree. However, a simpler solution (and obviously more in the spirit of MTT) would be to associate pairs of strings (i.e., the DMorphR, or linearized sequences of nodes), rather than just single strings, with nodes of the dependency tree. This approach is inspired by a generalization of CFG, called Head Grammar (HG) (Pollard, 1984), which has been shown to be formally equivalent to TAG (Weir et al., 1986). Basically, a HG provides a dependency tree and rules how to compute the final string (or "yield"). This string is computed bottom-up; with each node is associated a list of two string segments. As we go up the dependency tree, we compute the yield for each new node, based on the yields of its daughter nodes. The segments can be shifted around according to certain rules, and new terminal symbols added, but the segments may not be broken up. We see that this is exactly how the Surface-Syntactic Rules (SSyntR) of MTT operate (see, e.g., (Mel'čuk, 1967)), except that in the case of HG there are two strings instead of one.

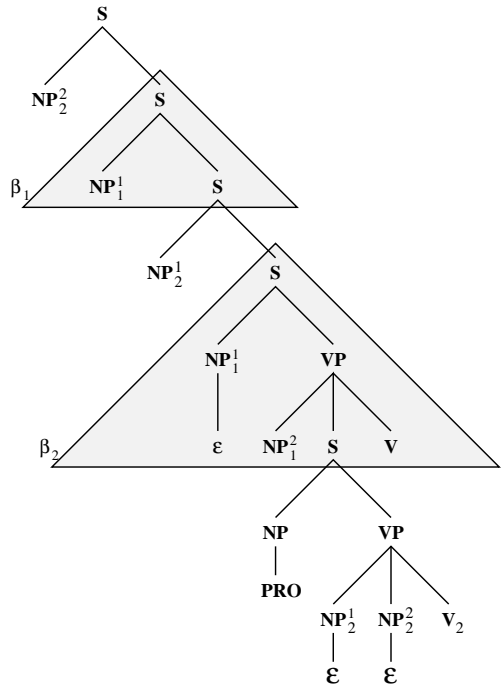

Figure 16: The Derived Tree for Sentence (6b)

Our proposal can best be illustrated by giving two SSynt rules (hopefully in the spirit of (Mel'čuk and Pertsov, 1987)) in which we use two-part strings to deal with the Dutch cross-serial dependencies (Figure 17). A dependency relation is now linearized not as one string, but as two, which are represented as a pair, separated by commas (e.g., *Y1, Y2*). Syntagm 2 takes care of the most deeply embedded clause: the verb X is put in the second segment, while all the overt nominal arguments are put in the first segment. Syntagm 3 applies when the most deeply embedded verb has no dependents at all. Syntagm 1 applies to verbs that subcategorize for clauses. The DMorphR associated with the embedded clause, Y, is in two segments, called Y1 and Y2. The governing verb of Y, X, is added to the left of Y2. Any nominal arguments (the subject or object) of X are added to the left of Y1. As an example, consider Dutch sentence (5) discussed previously. We give the SSyntR in Figure 18[12]. The noun phrase rooted in *kinderen* is of course linearized as *de + kinderen*. The clause rooted in in *zwemmen* has no dependents (verbal or nominal). Therefore, syntagm 3 applies, and we get a DMor-phR consisting of two strings, — and *zwemmen*. We then need to linearize the clause rooted in *leren*. Since *leren* does have a verbal dependent (namely *zwemmen*), syntagm 1 applies. We have Y1 = — (the empty string) and Y2 = *zwemmen*. The verb *leren* is added to the left of Y2. Furthermore, the condition in syntagm 1 specifies that the nominal arguments of *leren* must precede Y1. We therefore obtain:

(7) DMorphR for subtree rooted in *zwemmen*:

---

[12]The exact details, i particular the arc labels, are not of interest here. We also omit all features in the both the syntactic and morphological representations.

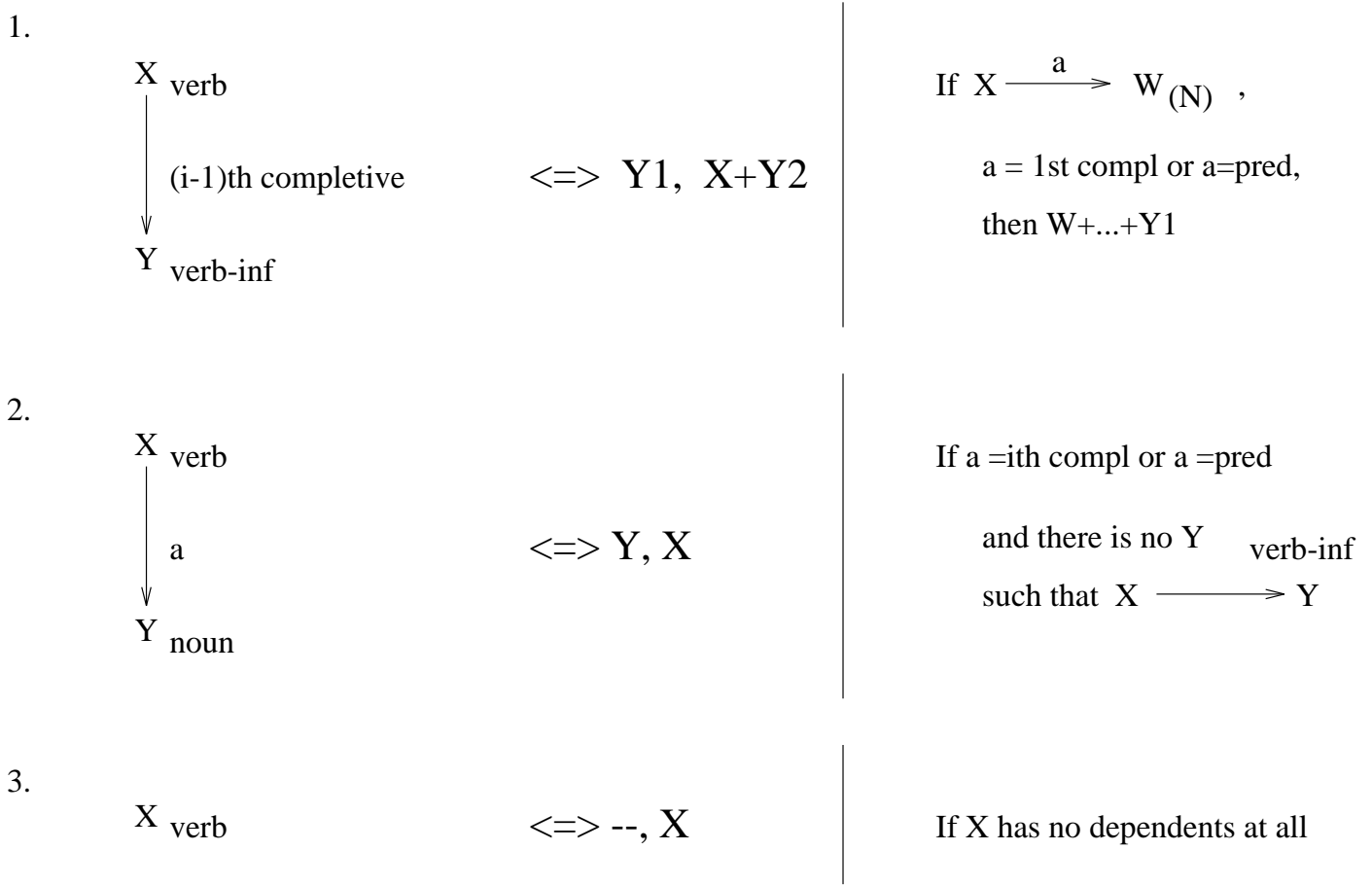

Figure 17: Syntagm for Dutch Embedded Clauses

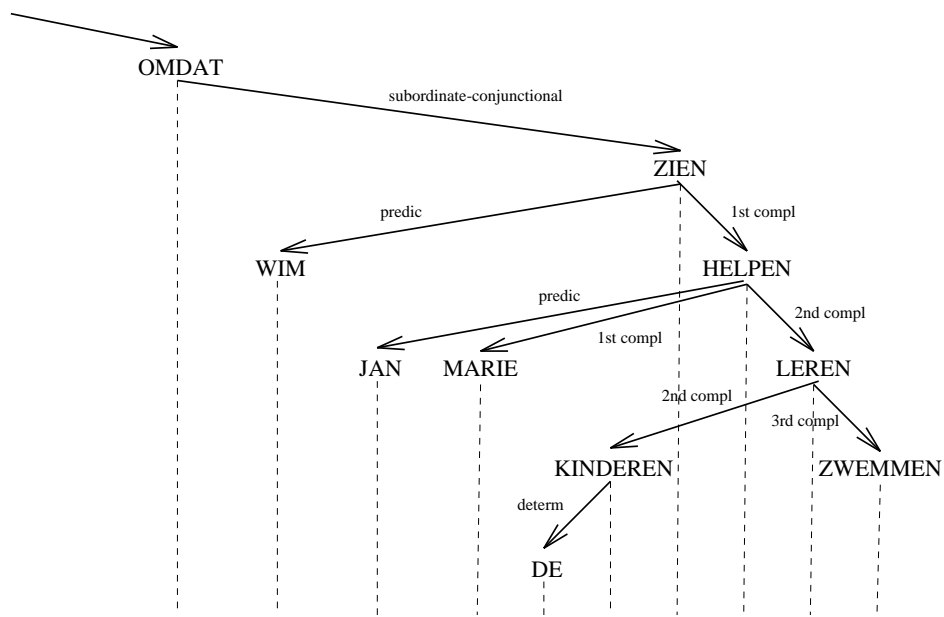

Figure 18: SSyntR for Sentence (5)

    de + kinderen,    leren + zwemmen

Now consider the subtree rooted in *helpen*. Again, syntagm 1 applies, this time with Y1 = *de + kinderen* and Y2 = *leren + zwemmen*. Again, the head verb is added to the left of Y2, while its nominal arguments are added to the left of Y1. We obtain:

(8) DMorphR for subtree rooted in *helpen*:

    Jan + Marie + de + kinderen,    helpen + leren + zwemmen

Finally, we apply syntagm 1 one more time for verb *zien*, and then the syntagm for the SUBORDINATE-CONJUNCTIONAL SSyntRel. This latter syntagm, not given here, will append the two parts of the DMorphR of its dependent *zien* node and append the *omdat* node, giving us the desired result:

(9) DMorphR for subtree rooted in *omdat*:

    omdat + Wim + Jan + Marie + de + kinderen + zien + helpen + leren + zwemmen

Thus, we do not need to have recourse to global rules: the word-order of the sentence is fixed in syntagms, despite the existence of unbounded deep non-projectivity. We can deal with embedded *wh*-words in a similar manner; for space limitations, we refrain from giving the details here.

Note that our proposal does not replace the notion of syntagm as defined in the Surface Syntactic-Component of MTT. Instead, it extends it, and it does so only in those cases where non-projectivity occurs (the other syntagms need not be changed). What is replaced is the notion of global ordering rules to handle cases such as English *wh*-movement and Dutch verb raising.

## 4 Conclusion

In this paper, we have argued that the crucial difference between a CFG-based analysis and a DG-based analysis is that the latter, but not the former, can be lexicon-based. We have described TAG, a phrase-structure grammar which can be lexicalized, and we have shown some similarities in linguistic analyses expressed in DGs and in TAGs. In considering non-projective word order phenomena, we have shown that two important results can be transferred form the TAG analysis to the MTT analysis: first, we can give upper bounds on processing complexity for specific constructions; second, we do not need to have two types of word-order rules, syntagm-based rules and global rules. Instead, if we extend the definition of a syntagm, all rules can be expressed locally.